# Investigating Group Distributionally Robust Optimization for Deep Imbalanced Learning: A Case Study of Binary Tabular Data Classification

Ismail. B. Mustapha[1], Shafaatunnur Hasan[2], Hatem S Y Nabbus[3], Mohamed Mostafa Ali Montaser[4], Sunday Olusanya Olatunji[5], Siti Maryam Shamsuddin[6]

Computer Science Department-School of Computing, Universiti Teknologi Malaysia, 81310 Skudai, Johor, Malaysia[1, 2, 3, 4, 6]
Higher Institute for Science and Technology, Al_shomokh – Tripoli, Libya[4]
Department of Computer Science-College of Computer Science and Information Technology, Imam Abdulrahman Bin Faisal University, Dammam, Saudi Arabia[5]

*Abstract*—One of the most studied machine learning challenges that recent studies have shown the susceptibility of deep neural networks to is the class imbalance problem. While concerted research efforts in this direction have been notable in recent years, findings have shown that the canonical learning objective, empirical risk minimization (ERM), is unable to achieve optimal imbalance learning in deep neural networks given its bias to the majority class. An alternative learning objective, group distributionally robust optimization (gDRO), is investigated in this study for imbalance learning, focusing on tabular imbalanced data as against image data that has dominated deep imbalance learning research. Contrary to minimizing average per instance loss as in ERM, gDRO seeks to minimize the worst group loss over the training data. Experimental findings in comparison with ERM and classical imbalance methods using four popularly used evaluation metrics in imbalance learning across several benchmark imbalance binary tabular data of varying imbalance ratios reveal impressive performance of gDRO, outperforming other compared methods in terms of g-mean and roc-auc.

*Keywords—Class imbalance; deep neural networks; tabular data; empirical risk minimization; group distributionally robust optimization*

## I. INTRODUCTION

Owing to increased data availability, novel learning architectures and accessibility to commodity computational hardware devices, deep neural networks (DNNs) have become the de facto tool for a wide range of machine learning (ML) tasks in recent times; leading to state-of-the-art performance in several computer vision, natural language processing and speech recognition tasks. DNNs are characterized by several layers of hidden units that enable learning of useful representations of a given data for improved model performance [1, 2]. This alleviates the need for domain experts and hand-engineered features, a common prerequisite for traditional ML methods.

A pervasive problem that has plagued traditional ML methods in the last couple of decades which DNNs are not immune to is the class imbalance problem [3-6]. This problem, also termed long-tailed data distribution problem in computer vision, occurs when the distribution of the constituent classes of a training data is highly disproportionate such that one or more classes have significantly larger number of training samples (majority class(es)) than other(s) (minority class(es)). Given that most ML methods are built to minimize the overall classification error with the assumption that each sample contributes equally, the learning algorithms tend to be bias towards the majority class; thus, resulting in partial or total disregard of the discriminative information of the minority classes by the learning algorithm. What makes this problem even more interesting is that, in most cases, the minority classes are often the classes of interest. Several manifestations of this problem abound in many real-life application domains of ML like medical diagnosis [7, 8], fraud detection [9-11], flight delay prediction [12, 13] amongst others.

The knowledge that learning from imbalance data negatively impacts the performance of DNN has resulted a marked increase in research on deep learning-based approaches to tackle the problem in recent years, with findings showing that traditional approaches to addressing imbalance problems can be successfully extended to DNN [3, 5]. Thus, many deep learning studies have addressed the imbalance problem at the data level mainly by data resampling [14-16] while some have done so at the classifier level, largely through cost sensitive learning approaches [17-22]. Oversampling and undersampling are two common data resampling approaches used in DNN. However, the susceptibility of the former to noise and overfitting due to added samples [23] as well as the characteristic loss of valuable information peculiar with the latter [3] remain major drawbacks of this category of imbalance methods. On the other hand, the core idea behind the cost sensitive methods is to assign different misclassification cost/weights to the training samples to scale up/down the misclassification errors depending on the class they belong [17, 24]. While there are several implementations of this method, the most commonly used cost sensitive approach in imbalanced deep learning research is reweighting [20, 25], where weights are assigned to different class samples based on either the inverse of the class frequencies [20, 26, 27] or their square root [28]. Despite its widespread adoption in DNN, reweighting methods have been found to be unstable in severely imbalanced cases; yielding poor performance that compromise the performance of the majority class [20, 23, 29, 30]. Inspired by the drawbacks of the commonly used methods, this study seeks to address the imbalance problem from a different perspective, through the learning objective.

The canonical training objective in DNN is the empirical risk minimization (ERM) which entails minimizing the average per sample training loss over the entire training data [31]. This training objective has the capacity to fit a given training data perfectly and still produce impressive accuracy on an unseen test data [32]. However, training a DNN using such objective on an imbalanced data has been shown to be bias to the majority class samples despite fitting the training data perfectly in most cases [32, 33]. The trained DNN is unable to generalize the learnt representations to the minority class samples at inference/test phase. In contrast to ERM, this study explores minimizing the maximum between the majority and minority class losses for improved imbalance learning. This is analogous to distributionally robust optimization (DRO) which seeks to minimize the expected loss over possible test distributions that the model is expected to perform well on [34, 35]. Specifically, group DRO (gDRO) proposed in [32] is investigated in this study in the context of classical class imbalance problem in DNN where the training and test data are similarly imbalanced. Rather than seeking reduction in the generalization gap between the training and test accuracies of the worst group, the performance of gDRO on binary imbalance datasets of varying imbalance ratios is investigated in comparison to popular imbalance methods in DNN. The performance of these methods is compared using four popular evaluation metrics in imbalance learning.

The rest of this article is outlined as follows: Section II provides the requisite background on ERM and gDRO followed by Section III which contains the methodology as it relates to the benchmark datasets, selected imbalance methods, DNN architecture and other experimental settings. The experimental results and discussions are presented and discussed in Section IV and Section V respectively before concluding in Section VI.

## II. THEORETICAL BACKGROUND ON ERM AND gDRO

Given a training data $\mathcal{D}_t = \{(x_i, y_i)\}_{i=1}^{N}$, where $y \in \mathcal{Y}$ and $x \in \mathcal{X}$ represent the target labels and the input features respectively, $f_\theta$ is a prediction function parameterized by $\theta$ that learns to correctly map each input feature $x_i$ to the corresponding output label $y_i$. The aim is to find the set of parameters $\theta$ that minimize the risk in (1).

$$\min_\theta \mathbb{E}_{(x,y) \sim \mathcal{D}_t}[\ell(y, f_\theta(x))] \quad (1)$$

where $\ell$ and $f_\theta(.)$ stand for the loss function and predicted output respectively. Equation (1) is approximated using the training set, $\mathcal{D}_t$, as in (2). This training objective is known as empirical risk minimization (ERM). In other words, the ERM aims to minimize the average per instance training loss.

$$\text{ERM} = \min_\theta \frac{1}{N} \sum_{i=0}^{N} \ell(y_i, \{f_\theta(x_i)\}_{c=1}^{C}) \quad (2)$$

A popular $\ell$ used in deep learning is the cross-entropy loss:

$$\ell(y_i, \{f_\theta(x_i)\}_{c=1}^{C}) = -\log p(y_i|x_i; \theta)$$

$$= -\log\left(\frac{\exp(f_\theta(x_i))}{\sum_{c=1}^{C} \exp(f_\theta(x_c))}\right) \quad (3)$$

The pseudocode for ERM is shown in Algorithm 1 below. In a classification problem, each output label $y_i$ belongs to one of $C$ classes and the aim of $f_\theta(.)$ is to ensure that each input $x_i$ is classified to the correct class $c \in \{1, ..., C\}$.

---

**Algorithm 1:** Empirical Risk Minimization Training

**Input:** *Training Data* $\mathcal{D}_t = \{(x_i, y_i)\}_{i=1}^{N}$; *no of Epochs, E; Batch size, b; Learning rate* $\eta$
**Output:** *Trained Model*
*Initialise model parameters* $\theta_0$
**for** *t = 1, 2...., E* **do**
 *randomly split* $\mathcal{D}_t$ *into* $n$ *equal-sized minibatches;* $|B| = b$
 **for** $B \in \{1..., n\}$ **do**
  *perform forward pass for model* $f_\theta$
  $\nabla_\theta L(\theta_t) \leftarrow \frac{1}{|B|} \sum_{i=0}^{b} \nabla_\theta L(y_i, f_\theta(x_i))$  #Compute loss and perform gradient step
  $\theta_{t+1} \leftarrow w_t - \eta. \nabla_\theta L(w_t)$  #Update model parameters in backward pass
 **end**
**end**

---

Unlike ERM, where the average per sample loss over the entire training data is minimized, gDRO minimizes the worst group error. Thus, gDRO presumes group annotations over the training data i.e., every training sample is a triplet $\{(x_1, y_1, g_1), ... (x_n, y_n, g_n)\}$ where the $g_n$ stands for the group annotation of the $nth$ sample. On the contrary, no group annotations based on spurious correlations are assumed in this study. Rather, a typical case of class imbalance where samples belonging to the minority class are fewer than those of the majority class is the focus of this study. Thus, number of samples belonging to each class is denoted as $N_c$. Hence, instead of (2), (4) is used to update the DNN in gDRO in this study; where $\frac{1}{\sqrt{N_c}}$ is for the group adjustment as in [32].

$$\text{gDRO} = \min_\theta \max_{c \in C} \left\{ \frac{1}{N_c} \sum_{i=1}^{N_c} \ell(y_i, f_\theta(x_i)) + \frac{1}{\sqrt{N_c}} \right\} \quad (4)$$

The pseudocode for group DRO training is presented in Algorithm 2.

## III. MATERIALS AND METHODS

### A. Benchmark Datasets

Nineteen (19) carefully selected binary class benchmark imbalanced datasets from Keel[1] and UCI[2] data repositories are used in this study. Details such as the sample size, number of features, fraction of majority and minority samples in percentage are presented in Table I. The degree of imbalance in each dataset is indicated by the imbalance ratio (IR) which is the ratio of the majority to minority samples size. Likewise, the degree of complexity of each dataset is shown using mean silhouette coefficient of its samples (S.Coeff) [36]. The S.Coeff values ranges between -1 and +1. Values around zero indicate overlapping class clusters, whereas values close to +1 and -1

---
[1] https://sci2s.ugr.es/keel/index.php
[2] https://archive.ics.uci.edu/ml/datasets.php

indicate well separated and highly overlapped class clusters respectively.

---

**Algorithm 2:** Group Distributionally Robust Optimization Training

**Input:** *Training Data* $\mathcal{D}_t = \{(x_i, y_i)\}_{i=1}^N$; *classes,* $c \in \{1, ..., C\}$; *no of Epochs, E; Batch size, b*

**Output:** *Trained Model*

*Initialize parameters of model* $f_\theta$
**for** *t = 1, 2...., E* **do**
   *randomly split* $\mathcal{D}_t$ *into* $n$ *equal-sized minibatches;* $|B| = b$
   **for** B ∈ {1..., n} **do**
     *perform forward pass for model* $f_\theta$
     **for** *c = {1..., C}* **do**
       $L_c(\theta_t) \leftarrow \frac{1}{N_c}\sum_{i=1}^{N_c} \ell(y_i, f_\theta(x_i)) + \frac{1}{\sqrt{N_c}}$ # *compute loss for each class*
     **end**
     $\nabla_\theta L(\theta_t) \leftarrow \nabla_\theta\{max(L_c(\theta_t)_{c=1}^C)\}$ # *perform gradient step with worst group*
     $\theta_{t+1} \leftarrow \theta_t - \eta \cdot \nabla_\theta L(\theta_t)$    # u*pdate model parameters in backward pass*
   **end**
**end**

---

### B. Methods for Handling Class Imbalance

In addition to ERM and gDRO, experiments were also carried out using four classical imbalance methods and compared. These methods were chosen to cover commonly used imbalanced methods in DNN research and their hybrid.

- Random Oversampling (ROS)
- Random Undersampling (RUS)
- Cost sensitive reweighting (COST): The weights assigned to majority and minority class samples are determined by the inverse of $N_c$; where $N_c$ is the number of samples belonging to class $c$.
- Hybrid of random undersampling and oversampling (RUSROS): This involves initially randomly undersampling the majority class by 50% before randomly oversampling the minority class samples till it equals the majority class size.

Overall, the performance of six methods (ERM, gDRO, ROS, RUS, COST and RUSROS) on the imbalance datasets are compared in this study.

### C. DNN Architecture

Unlike convolutional neural networks (CNN), where a wide range of benchmark architectures are available [37], determining an appropriate DNN architecture for tabular data is nontrivial due to the sparsity of representative works addressing pertinent issues such as the ideal network depth and width as well as the best activation functions for this class of models. While in recent years several novel architectures have been proposed in representative studies [38-41], no single method provides a reliable performance across multiple tasks. Hence, deep fully connected otherwise known as deep multilayer perceptron remains the quintessential baseline architecture for modelling structured data [41] and thus, used in this study. Besides, deep fully connected neural networks are natural fit for imbalanced data with capability of yielding impressive results when the hyperparameters are optimized [42].

TABLE I. BENCHMARK DATASETS

| Data | # Samples | # Features | % Maj Class | % Min Class | IR | S.Coeff |
|---|---|---|---|---|---|---|
| **abalone19** | 4174 | 8 | 99.23 | 0.77 | 129.44 | -0.021 |
| **protein_homo** | 145751 | 74 | 99.11 | 0.89 | 111.46 | 0.556 |
| **mamography** | 11183 | 6 | 97.68 | 2.32 | 42.01 | 0.45 |
| **ozone_level** | 2536 | 72 | 97.12 | 2.88 | 33.74 | -0.049 |
| **wine_quality** | 4898 | 11 | 96.26 | 3.74 | 25.77 | 0.146 |
| **oil** | 937 | 49 | 95.62 | 4.38 | 21.85 | 0.084 |
| **abalone** | 731 | 8 | 94.25 | 5.75 | 16.4 | 0.107 |
| **glass4** | 214 | 9 | 93.93 | 6.07 | 15.46 | 0.363 |
| **covertype** | 38501 | 54 | 92.87 | 7.13 | 13.02 | 0.114 |
| **vowel0** | 988 | 13 | 90.89 | 9.11 | 9.98 | 0.166 |
| **satimage** | 6435 | 36 | 90.27 | 9.73 | 9.28 | -0.134 |
| **page-blocks0** | 5472 | 10 | 89.78 | 10.2 | 8.79 | 0.505 |
| **ecoli3** | 336 | 7 | 89.58 | 10.4 | 8.6 | 0.126 |
| **segment0** | 2308 | 19 | 85.75 | 14.3 | 6.02 | -0.063 |
| **yeast4** | 1484 | 8 | 83.56 | 16.4 | 5.08 | 0.037 |
| **vehicle0** | 846 | 18 | 76.48 | 23.5 | 3.25 | 0.065 |
| **haberman** | 306 | 3 | 73.53 | 26.5 | 2.78 | 0.069 |
| **phoneme** | 5404 | 5 | 70.65 | 29.4 | 2.41 | 0.087 |
| **pima** | 768 | 8 | 65.1 | 34.9 | 1.87 | 0.092 |

ReLU activation function is widely used in deep learning class imbalance research [14, 43], hence the same is adopted in the DNN model used in this study. Batch normalization and 0.5 dropout rate are applied after each ReLU activation of each hidden layer to avoid overfitting. Likewise, to optimize a DNN model for each dataset, representative imbalance studies have often resulted to grid search for hyperparameters optimization [14, 44, 45]. Similarly, 80% of each imbalanced data was used for hyperparameter optimization via grid search. Only the depth and width of each DNN model are optimized as in [14]. A network width of 50 neurons per hidden layer was found to be sufficient for the models to overfit the data after experimenting with widths of 512, 300, 100, 50 and 32 neurons respectively. The depth of these models was optimized starting with a depth of two (i.e., 2-hidden layers) and varying it up to six. The optimal DNN architecture for each imbalanced dataset was determined via the best mean AUC over 5-fold cross validation [42]. The AUC results for optimal number of layers and architecture for each dataset is presented in Table IV of the appendix.

*D. Experimental Setup*

10-fold cross validation approach was employed for model training and validation for each combination of dataset, model and imbalance method. All DNN models were trained to stop early if there are no improvement in the validation error after 10 successive epochs. Adam stochastic optimization with learning rate of 0.001 was used in model training. Each model weights were randomly initialized with uniform distribution and Xavier variance [46] with zero bias before training with batch stochastic gradient descent. The minibatch sizes were set to range from 1/32 to 1/100 of each respective dataset sample sizes. The training procedure is illustrated in Fig. 1.

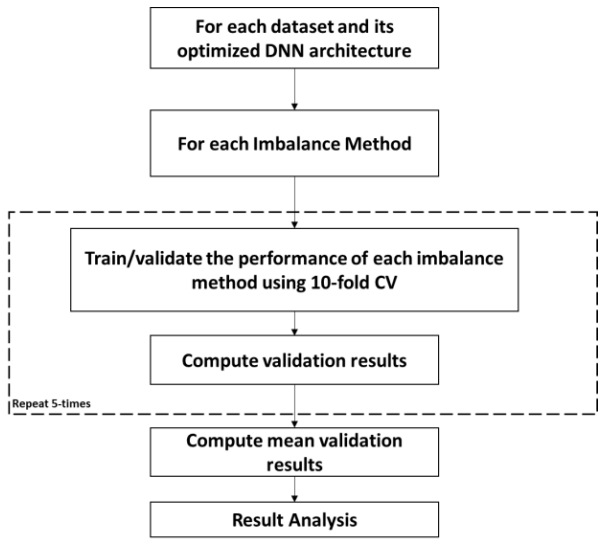

Fig. 1. Training procedure

To further ensure credibility of our findings and handle associated inconsistencies resulting from randomization and stochastic nature of the training process, the experiment for each combination was repeated five times with varying random seeds; resulting in different initial parameters for each repetition [47]. The mean validation results across the repetitions are reported. In all, 5,700 experiments were conducted. Given the large number of models that is required to be trained, a commodity hardware GPU, Nvidia Geforce GTX 960M, on a core i5 Dell Inspiron 7559 machine was leveraged to speed up the experiments. All experiments were implemented in Pytorch deep learning framework [48].

*E. Evaluation Metrics*

The inadequacy of accuracy and error rate as measures of classification performance of imbalance datasets is well documented in the literature [49]. Thus, four complementary evaluation metrics that have been used in imbalance learning research have been adopted in the study. Each of this metrics is described in what follows. Note that FP, FN, TP and TN are false positive, false negative, true positive and true negative respectively.

- Receiver Operating Characteristic Area Under the Curve (ROC-AUC): The ROC is a plot of the true positive rate ($\frac{TP}{TP+FP}$) against the false positive rate ($\frac{FP}{FP+FN}$) across all possible discrimination thresholds. From this plot, the AUC which is the area under the receiver operating characteristic (ROC) curve can be calculated and used as a performance measure of classification model.

- Precision-Recall AUC curve (PR-AUC), perhaps inspired by the ROC-AUC, is a plot of precision on the y-axis ($\frac{TP}{TP+FP}$) against recall ($\frac{TP}{TP+FN}$) on the x-axis. The area under the PR-curve is also used as a measure of the performance of binary classification models. The AUC implementation used in this work is calculated using the trapezoidal rule.

- F1-Measure, a widely used evaluation metric, is another measure of evaluation used in this study. It is the harmonic mean of precision and recall (i.e., $\frac{(1+\beta^2) \times Recall \times Precision}{\beta^2 \times Recall + Precision}$). The beta ($\beta$) parameter shows the trade-off between precision and recall. Our interest is to detect both majority and minority classes with equal preference, hence, the $\beta$ parameter is set to 1.

- Geometric mean (g-mean) is the final evaluation metric used to evaluate the models. In case of binary classification, g-mean is the squared-root of the product of recall and true negative rate (TNR) ($\sqrt{\frac{TP}{TP+FN} \times \frac{TN}{TN+FP}}$).

Additionally, the Friedman test [50] is also used in this study to detect differences in the experimental results across multiple attempts, when the normality assumption may not hold. Thus, this test is used to reject the null hypothesis that the compared methods produce similar performance across the different datasets and DNNs in comparison to their mean rankings. Then, as recommended in [51], pairwise posthoc analysis using Wilcoxon signed-rank test [52] with Holm's alpha (0.05) correction [53] was used for comparison. A visualization of the comparison is presented using a critical difference diagram [54].

IV. RESULTS

Multiple benchmark datasets enabled a fair comparative analysis of ERM, gDRO and the imbalance methods across a wide range of imbalance ratios. The mean (±standard deviation) of the validation performance of each method across the respective evaluation metrics for each dataset over five repetitions of the experiment (as shown in Fig. 1) are presented in Fig 2. The overall average performance of each method in addition to the number of times each method ranks first for each evaluation metric is also presented in Table II. Similarly, a bar plot showing the average ranking of each method per dataset is presented in Fig. 3.

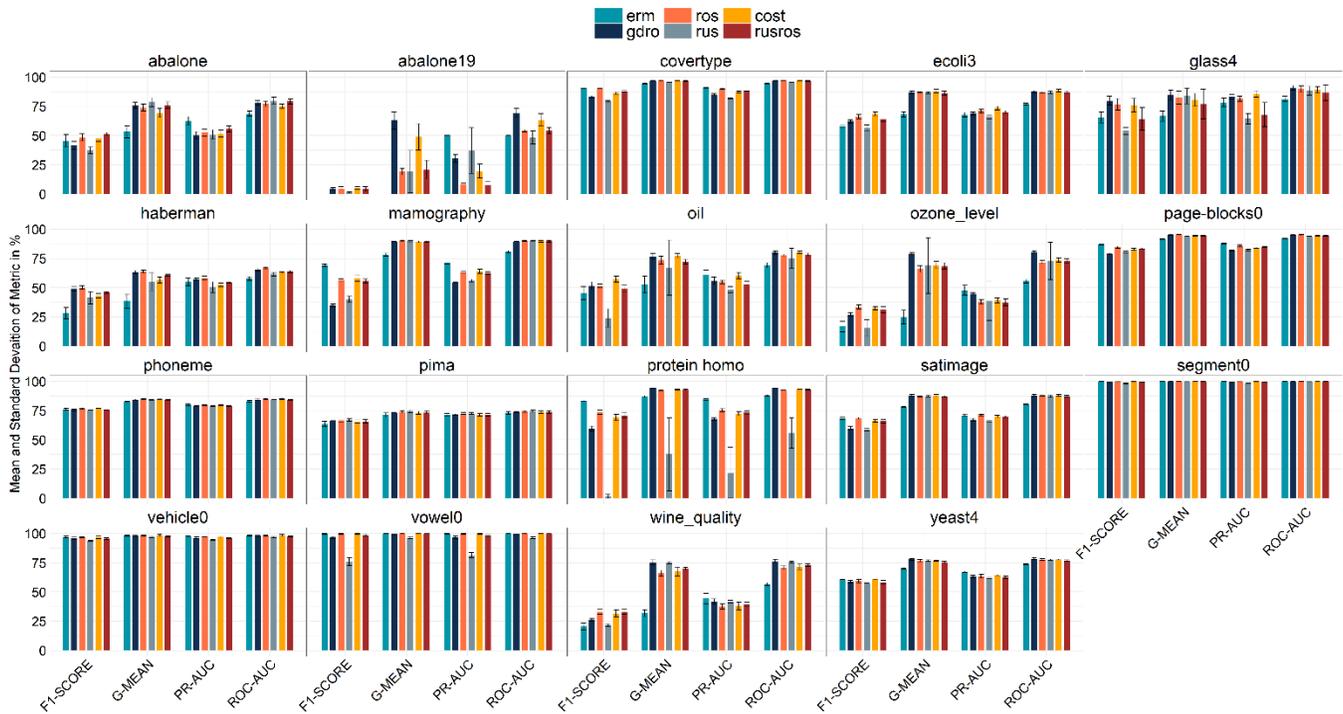

Fig. 2. Mean (±standard deviation) results of comparative methods

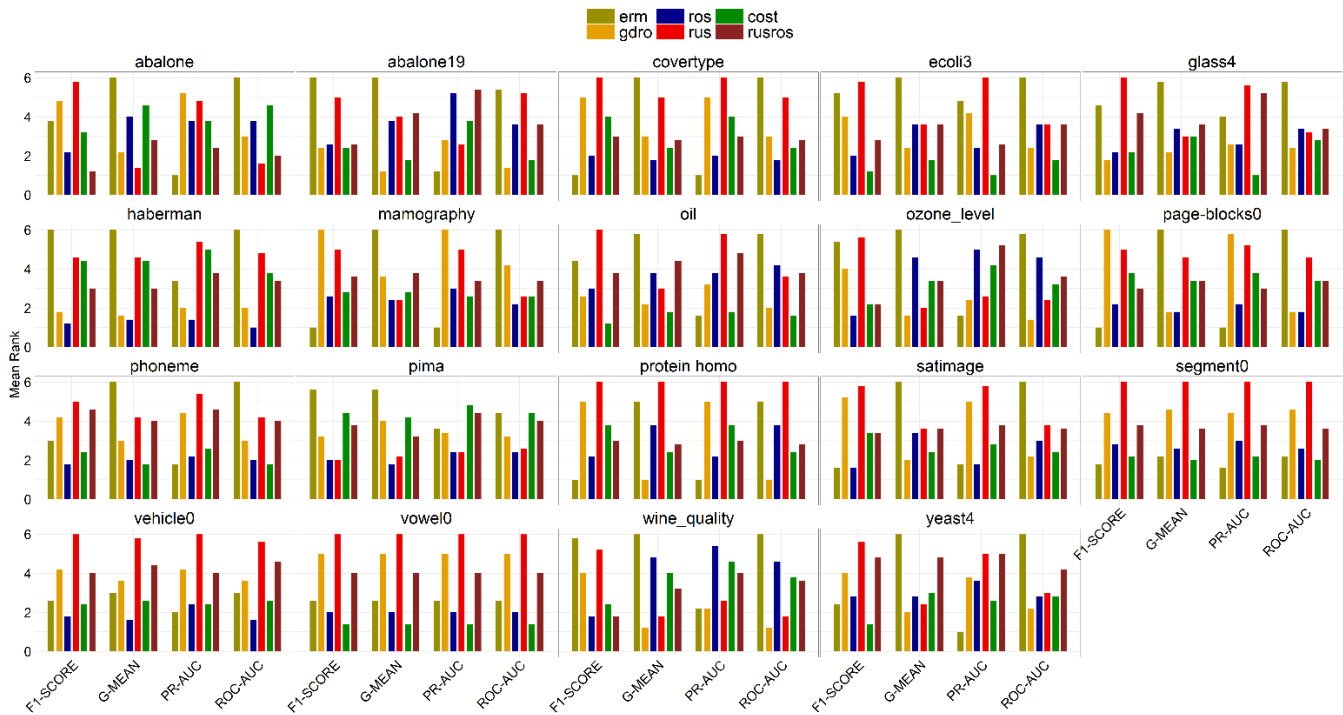

Fig. 3. Mean ranking of the experimental results (lower the better)

*A. Results based on F1-Score*

As, depicted in Fig. 2, the performance of gDRO in comparison to ERM regarding the top six imbalanced benchmarks (IR of 129.44 to 21.85) shows that gDRO mostly produces better average f1-score for those with overlapping classes (abalone19, ozone level, wine_quality and oil) i.e., gDRO identifies minority samples better on complex imbalanced datasets- whereas ERM performed better on those with better class separability (protein homo and mammography). The general f1-score on the abalone19 benchmark is notably very poor across the compared methods, a likely explanation for this is lack of data/information (given a meagre minority sample size of 32) worsened by class overlap.

Likewise, the superior performance of gDRO over ERM on this benchmark suggests the suitability of the former in extreme imbalance cases where data is lacking. As for the remaining datasets in decreasing order of imbalance ratio, ERM produced better average f1-score than gDRO in all but glass4, ecoli3 and haberman datasets; the three smallest datasets in the benchmarks and of varying degree of overlap. In comparison to the classical imbalance methods, while mostly outperforming RUS on mammography, page-block0 and pima, gDRO mostly produced lower average f1-score than, at least, one of ROS, COST or RUSROS except on abalone19 and glass4 datasets where gDRO outperformed other methods. As shown in Table II, ROS produced the best average overall f1-score across the benchmarks, producing the best score seven times in the process. On the other hand, RUS produced the lowest overall f1-score and only managed to produce the best score once (which was jointly with ROS on the least imbalance dataset, pima).

### B. Results based on G-Mean

As shown in Fig. 2, gDRO produced better average g-mean scores than ERM across all but the segment0, vehicle0 and vowel0 datasets where they both performed similarly or marginally better performance by ERM. It should be noted that the performance of the model on these datasets is generally better than the remaining datasets. In comparison to the selected imbalance methods in terms of the top six imbalanced datasets (abalone19, protein homo, mammography ozone level, wine_quality and oil), gDRO yielded the best average g-mean on all but the mammography and oil datasets where ROS and RUS as well as COST respectively produced better performance. On the remaining datasets, as the imbalance ratio reduces and the minority to majority ratio increases, the performance of the classical imbalance methods generally become more comparable to gDRO which was only able to produce the best average g-mean on the glass4, satimage, page-block0 and yeast4 datasets. Table II shows that gDRO produced the best average overall g-mean of 84.28% across the benchmarks, while achieving the best score 8 times in the process. The implication of the impressive performance of gDRO in terms of g-mean is that it detects the minority samples with lower false positive and false negative rates. On the other hand, an overall average g-mean score of 67.74% achieved by ERM makes it the least performing method in this regard.

### C. Results based on PR-AUC

As shown in Fig. 2, ERM outperforms gDRO and other classical imbalance methods based on average pr-auc on all but the ecoli3, glass4, haberman and pima datasets. Three of these datasets (ecoli3, glass4 and haberman) have the least number of samples while the relatively larger pima dataset is the least imbalance amongst the considered benchmarks. Despite achieving better average pr-auc than ERM on these datasets, the performance of gDRO still remains inferior to at least one of ROS, RUS, COST or RUSROS. Table II further shows that ERM produced the best overall average pr-auc of 73.07%, achieving the best performance 14 times across the different benchmarks. In contrast, RUS showed the lowest performance in this regard despite producing the best result once (jointly with ROS on the pima dataset).

TABLE II. AVERAGE PERFORMANCE ACROSS ALL DATASETS (AND THE NUMBER OF TIMES EACH METHOD RANKS FIRST)

| Metric | ERM | GDRO | ROS | RUS | COST | ROSRUS |
|---|---|---|---|---|---|---|
| F1-SCORE | 61.85 (6) | 60.57 (2) | 65.07 (7) | 51.54 (1) | 64.37 (5) | 63.19 (2) |
| G-MEAN | 67.74 (0) | 84.28 (8) | 80.36 (6) | 77.23 (2) | 81.68 (5) | 80.12 (0) |
| PR-AUC | 73.07 (14) | 68.16 (1) | 68.41 (3) | 62.83 (1) | 68.95 (3) | 66.92 (0) |
| ROC-AUC | 77.86 (0) | 85.55 (8) | 83.80 (4) | 81.10 (0) | 84.28 (0) | 83.58 (1) |

### D. Results based on ROC-AUC

The performance of gDRO as illustrated in Fig. 2 relative to ERM in terms of average roc-auc is similar to findings based on average g-mean as gDRO shows better average roc-auc scores than ERM across all but the segment0, vehicle0 and vowel0 datasets where they both performed similarly or ERM is marginally better. Comparison based on the top six imbalanced benchmarks also shows similar trend as gDRO produced the best average roc-auc on all but the mammography (where ROS, RUS, COST and RUSROS were better) and oil (where COST was better) datasets. Datasets with lower IR produced improved results that are comparable or relatively better than gDRO with these methods. However, gDRO produced the best average roc-auc on the glass4, satimage, page-block0 and yeast4 datasets. Further, Table II shows that gDRO produced the highest overall average roc-auc (85.55%), producing the results eight times across the datasets. On the other hand, ERM produced the lowest overall average performance based on ROC-AUC.

### E. Statistical Analysis

The Friedman's test results presented in Table III shows that the null hypothesis, namely, the imbalance methods produce similar performance across the different datasets is rejected. Hence, pairwise posthoc analysis using Wilcoxon signed-rank test is carried out to rank each method across the evaluation metrics as illustrated in Fig. 4. The figure shows a critical difference diagram of the imbalance methods for each evaluation metric where a thick horizontal line indicates group of imbalance methods (a clique) for which the difference in their performance is not statistically significant. For each metric, the mean performance of each method across the dataset is used for the statistical test. The figure shows that although for f1-score, g-mean, pr-auc and roc-auc, ROS, gDRO, ERM and gDRO respectively rank highest, each of these methods is not significantly better than at least two other methods. For instance, in terms of f1-score, ROS is not statistically different from COST and ERM whereas in term of pr-auc, no statistically significant difference exists between ERM, ROS, COST and gDRO.

TABLE III. RESULT OF FRIEDMAN'S TEST

| Metric | p-value |
|---|---|
| F1-Score | 0.00000 |
| G-Mean | 0.00000 |
| PR-AUC | 0.00000 |
| ROC-AUC | 0.00000 |

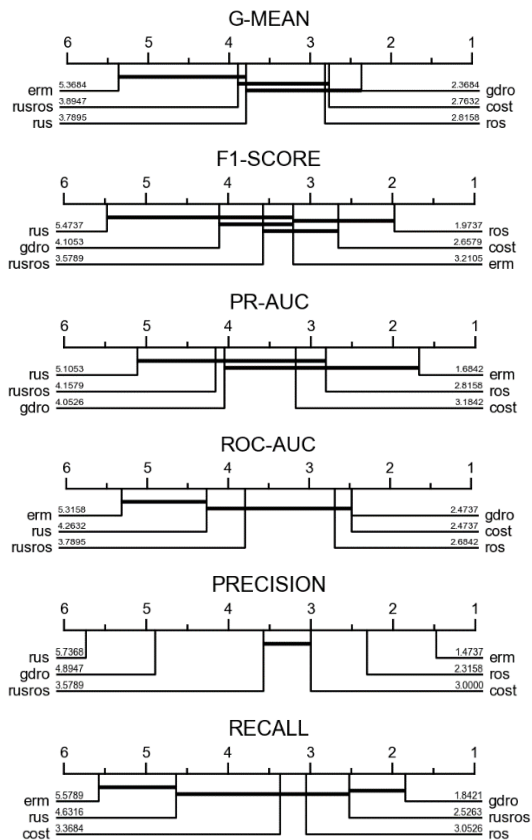

Fig. 4. Critical difference diagram of pairwise statistical difference comparison between imbalance methods

## V. DISCUSSION

The importance of this study cannot be overemphasized as it investigates gDRO for learning imbalance tabular data in DNN as against ERM which is the main learning objective in both balanced and imbalanced deep learning research. The empirical results of this study show that in terms of ROC-AUC which reflects a balanced assessment of the performance of the DNN models on both majority and minority samples across different thresholds, gDRO mostly outperforms ERM and most of the compared imbalance methods on benchmark datasets of varying imbalance ratios, sizes and complexities. This implies that deep imbalance learning via minimization of worst-case loss across classes can produce models that are more robust to both majority and minority class samples than ERM. Hence, gDRO is an ideal training objective in cases where both majority and minority classes of the imbalance data are equally important as it is known for its robustness to distributional shifts [32].

Following the generally held notion that ERM biases DNN models to the majority class when learning imbalance data [55], it would be expected that ERM perform poorly on the minority samples. However, it can be observed that ERM mostly outperform gDRO and other compared imbalance methods across the studied benchmarks in terms of pr-auc which mainly focuses on the performance of the DNN model on the minority samples under a range of thresholds. One likely explanation for this is that while ERM produced the best overall result on the minority samples over different thresholds, this does not necessarily mean it produced the best performance for a specific threshold value [56]. This explains why ERM is unable to replicate similar feat on metrics like f1-score and g-mean that are computed for specific threshold value.

Additionally, since pr-auc measures the area under the plot of precision against recall for different thresholds, another possible explanation could be that the pr-auc of ERM is dominated by precision. As hinted in Section E of IV, precision quantifies the number of correct positive (minority) predictions by dividing the number of correctly classified positive samples by the total number of correctly classified positive samples and negative (majority) samples that are incorrectly classified as positive. Compared to recall which quantifies the number of correct positive predictions from all positive samples, only a model that is bias to the majority samples is less likely to misclassify a majority sample as minority than misclassify a minority sample. In other words, ERM could produce the best pr-auc while being bias to the majority class samples if pr-auc is dominated by precision. The critical difference diagrams for precision and recall have been included in Fig. 4 to further elucidate this line of thought. See Fig. 5 and Fig. 6 of the appendix for more details of each method on each dataset in terms of precision and recall.

In relation to the classical imbalance methods, COST and ROS tend to perform similarly and better than RUS in most cases with COST showing superior performance in highly imbalance cases. However, the performance gains in terms of minority sample detection for these methods tend to come at the expense of some majority class samples [20]. On the other hand, RUSROS does not appear to have any major advantage that its constituent methods have not exhibited. Generally, the inferior performance of the imbalance methods on the highly imbalance benchmarks with some degree of class overlap like abalone19 and ozone_level compared to similarly imbalance ones like protein homo and mamography with more minority samples and better class separability underlines the impact of size and complexity of data in imbalance learning. Decreasing imbalance ratio tend to result in improved performance across the metrics.

It should however be noted that the adopted experimental design could have impacted the empirical results of this study. Particularly, in relation to reporting the mean scores of several repetitions of the experiments as against a single round as common in most DNN-based imbalance research [5]. Nevertheless, the adopted design has obvious benefits amongst which is an objective measure of the true model performance.

In sum, the choice of method for handling class imbalance in DNN models depends on several factors, including the severity of class imbalance, the size and complexity of the dataset, and the specific evaluation metric of interest. In some cases, ERM may be sufficient to achieve good performance on imbalanced datasets, while in other cases, gDRO or other methods may be necessary.

## VI. Conclusion

Deep imbalance learning has mainly focused on imbalance in computer vision related tasks. Likewise, empirical risk minimization (ERM) which entails minimizing the average per sample training loss over the entire training data has been shown in pertinent works to bias DNNs to the majority class when learning from imbalance data [31]. An alternative learning objective, group distributionally robust optimization (gDRO) is investigated in this study for imbalance learning with a focus on tabular data. The performance of gDRO in comparison with ERM and four classical imbalance resolution methods on several benchmark imbalance datasets of varying imbalance ratios are examined using four common metrics for evaluating class imbalance. Experimental findings show that while gDRO outperform other methods in terms of g-mean and ROC-AUC, whereas ERM and ROS rank highest in terms of pr-auc and f1-score respectively. Future research efforts will focus on the impact of pretraining on deep imbalance learning as well as gDRO for multiclass imbalance tabular data.

APPENDIX

TABLE IV. ROC-AUC SCORES FOR OPTIMAL ARCHITECTURE SELECTION FOR EACH DATASET

| Dataset | 2_lyrs | 3_lyrs | 4_lyrs | 5_lyrs | 6_lyrs |
|---|---|---|---|---|---|
| abalone19 | 50 | 50 | 50 | 50 | **50** |
| abalone | 66.33358 | 64.81243 | 71.23284 | **71.59815** | 66.15334 |
| covertype | 93.75338 | **94.1079** | 93.28036 | 93.94478 | 93.53545 |
| ecoli3 | 71.14201 | 64.17092 | 69.55867 | 71.43367 | **75.80867** |
| glass4 | 79.375 | 74.6875 | 74.6875 | 74.6875 | **84.6875** |
| haberman | **65.53534** | 61.76715 | 62.7362 | 63.29175 | 61.96581 |
| mamography | 74.85703 | 76.7389 | 77.4536 | **78.67885** | 78.64452 |
| oil | 69.72028 | 72.29672 | **72.50651** | 71.21878 | 61.21878 |
| ozone_level | **54.46271** | 50.76923 | 50 | 51.64122 | 52.2568 |
| page-blocks0 | 90.47406 | 90.94058 | 90.8244 | **91.70862** | 90.76801 |
| phoneme | 82.28141 | 83.61735 | 83.42348 | 83.15185 | **85.3148** |
| pima | 72.05325 | 72.35902 | 73.29717 | 72.05595 | **73.51786** |
| protein_homo | 85.52046 | 86.38106 | **88.44263** | 87.1977 | 87.19304 |
| satimage | 78.53431 | **80.78899** | 78.5861 | 79.28501 | 78.04006 |
| segment0 | **99.96845** | 99.77978 | 99.74823 | 99.5911 | 99.74823 |
| vehicle0 | 96.21446 | 95.72569 | 96.74197 | 95.51737 | **96.86146** |
| vowel0 | 99.93056 | 99.93056 | 99.21627 | 99.14683 | **100** |
| wine_quality | 54.41053 | **56.35424** | 54.12273 | 52.58129 | 52.51172 |
| yeast4 | 73.4827 | 73.3687 | 73.96551 | 72.27256 | **75.26621** |

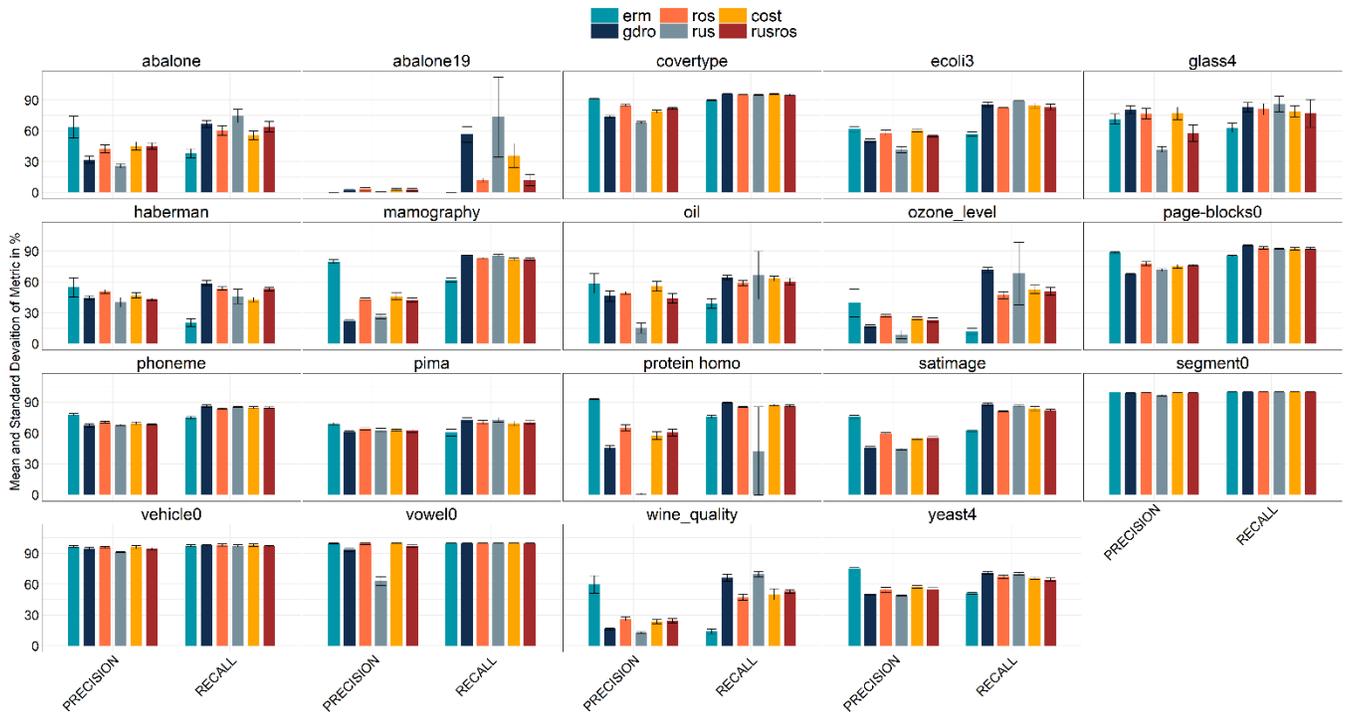

Fig. 5. Mean (±standard deviation) for precision and recall of comparative methods

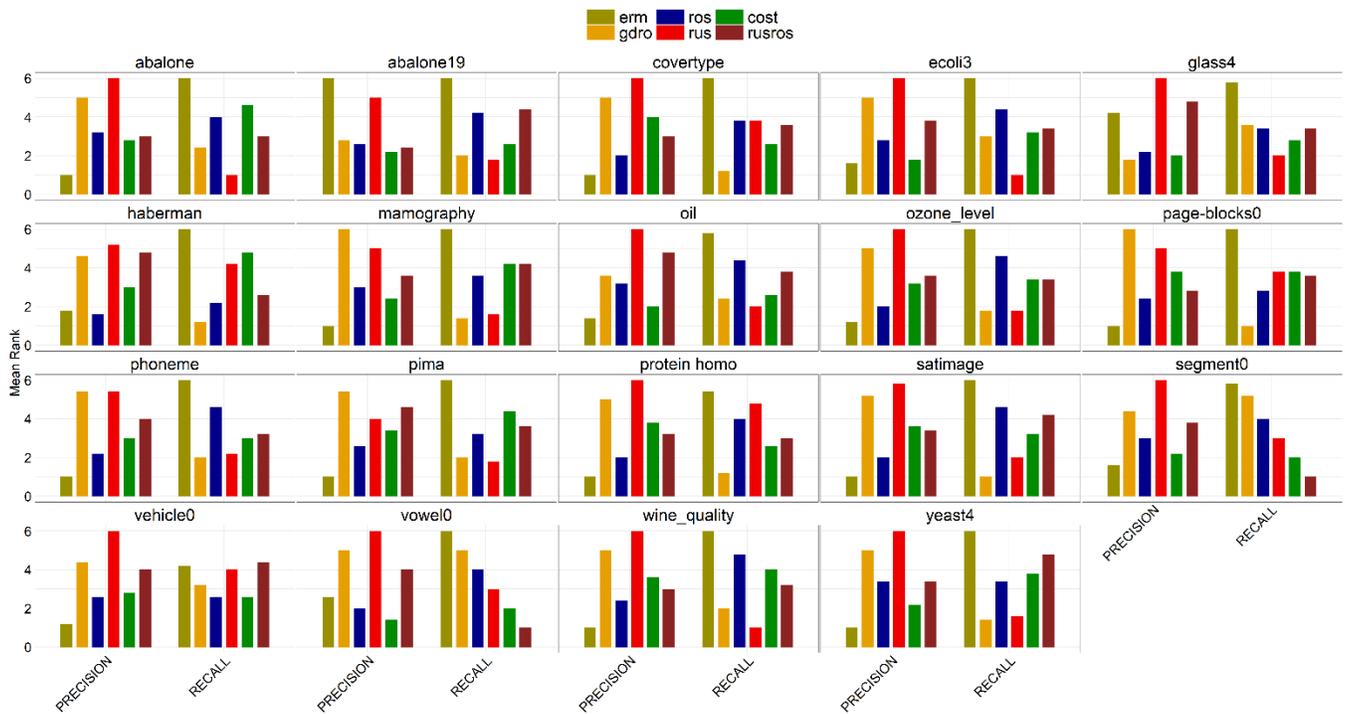

Fig. 6. Mean ranking precision and recall (lower the better)